# LGESynthNet: Controlled Scar Synthesis for Improved Scar Segmentation in Cardiac LGE-MRI Imaging

Athira J. Jacob[1,2], Puneet Sharma[1], and Daniel Rueckert[2,3]

[1] Digital Technology and Innovation, Siemens Healthineers, Princeton, NJ, USA
[2] AI in Healthcare and Medicine, Klinikum rechts der Isar, Technical University of Munich, Germany
[3] Department of Computing, Imperial College London, UK

**Abstract.** Segmentation of enhancement in LGE cardiac MRI is critical for diagnosing various ischemic and non-ischemic cardiomyopathies. However, creating pixel-level annotations for these images is challenging and labor-intensive, leading to limited availability of annotated data. Generative models, particularly diffusion models, offer promise for synthetic data generation, yet many rely on large training datasets and often struggle with fine-grained conditioning control, especially for small or localized features. We introduce LGESynthNet, a latent diffusion-based framework for controllable enhancement synthesis, enabling explicit control over size, location, and transmural extent. Formulated as inpainting using a ControlNet-based architecture, the model integrates: (a) a reward model for conditioning-specific supervision, (b) a captioning module for anatomically descriptive text prompts, and (c) a biomedical text encoder. Trained on just 429 images (79 patients), it produces realistic, anatomically coherent samples. A quality control filter selects outputs with high conditioning-fidelity, which when used for training augmentation, improve downstream segmentation and detection performance, by up-to 6 and 20 points respectively.

**Keywords:** Latent Diffusion Models · Image Synthesis · LGE MRI.

## 1 Introduction

Late Gadolinium Enhancement (LGE) cardiac MRI is the gold standard for assessment of myocardial viability, providing information on scar and fibrosis. However, accurate enhancement segmentation (referred to as LGE or scar segmentation henceforth)is challenging due to its subtle patterns, imaging artifacts, protocol variability, and the need for advanced expertise—leading to limited datasets for training deep learning (DL) models.

To mitigate data scarcity, recent work has explored generative models, especially diffusion models (DMs), for synthetic image generation [26, 11]. While DMs offer high image fidelity and controllability via spatial conditioning (e.g., masks, edge maps), they often require large datasets and struggle with conditioning fidelity—particularly for small or sparse structures like myocardial scars, which typically occupy <1% of the image. Lack of semantic masks over most of the image (e.g., outside the heart) further complicates generation by making the task under-constrained.

We propose LGESynthNet, a latent diffusion model (LDM) framework for generating LGE images with controllable scar morphology. Framed as an inpainting task, the model is trained on just 429 scar-positive LGE images from 79 patients. Built on ControlNet[28], our model incorporates (a) reward-based conditioning supervision, (b) anatomically descriptive caption generation, and (c) a biomedical domain-specific text encoder. We define evaluation metrics across image realism, conditioning consistency, and downstream utility, and demonstrate that augmenting segmentation models with LGESynthNet outputs improves Dice by upto 6 points and patient-level accuracy by upto 20 points.

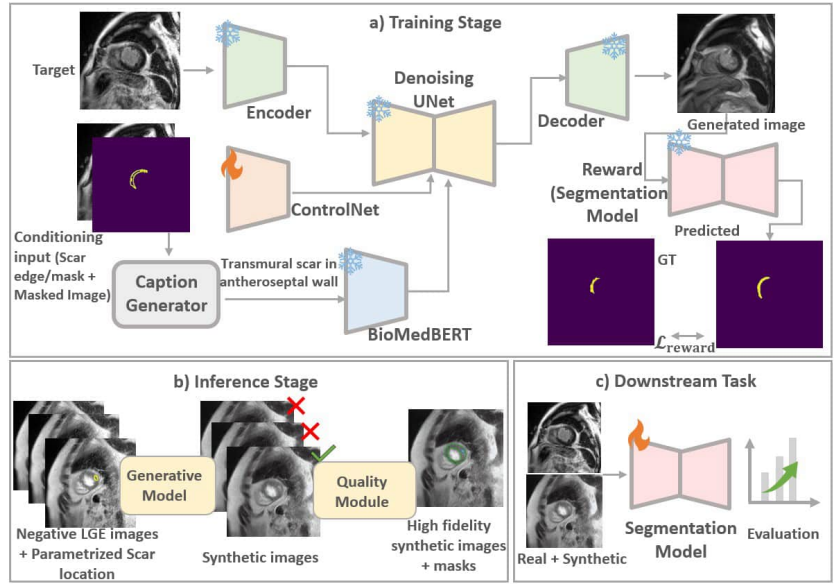

**Fig. 1.** Overview of the proposed method. During training, the model is conditioned on "masked" positive LGE images and intended scar boundaries. During inference, the model is conditioned on negative LGE image and the intended scar boundaries for the image. A quality control stage selects high quality synthetic images for use in the downstream task.

## 2 Related Work

**Semantic Image Synthesis.** GAN-based methods like pix2pix [8] and SPADE [21] have explored conditional image synthesis from semantic masks. More recently, Diffusion Models (DMs)[5] have demonstrated superior image quality via iterative denoising, with frameworks like ControlNet [28] and T2I-Adapter [19] leveraging prompts such as edge maps and masks. These models have been applied to medical imaging (retinal imaging [10, 14], CT [12, 25], MRI [12]) also improving downstream performance [12, 25, 24]. Diffusion models have generally outperformed GANs in fidelity and diversity [20], but often require large datasets and struggle with weak conditioning [17, 27] and fine-grained detail generation [25].

**LGE Analyses.** Quantitative assessment in LGE MRI lacks standardization [4], with methods like Full Width Half Maximum (FWHM)[6] and the n-standard deviation method[3] showing high variability. DL approaches [16, 29] show promising performance. The EMIDEC challenge [16] reported Dice scores between 0.27–0.71 and classification accuracies from 62–91%. Deng et al. [2] applied DMs for LGE synthesis but relied on paired multi-sequence inputs and lacked explicit control over enhancement properties. Ramzan et al. [22] use pixel-space DMs to synthesize LGE scar, achieving Dice upto 0.635. While LDMs have been shown to be more computationally efficient due to their learning in latent space, with equal or superior image quality, compared to pixel-space DMs [23], no studies explore their use for conditional LGE synthesis.

## 3 Methods

### 3.1 LGESynthNet

**Overview.** The architecture (Figure 1) consists of a latent diffusing U-Net, a ControlNet encoder [28], a text encoder, and a reward model. Generation is framed as inpainting, conditioned on (a) a scar edge map (via Canny detection of mask) and (b) an LGE image with the scar region masked with its mean intensity. The edge map guides location; the masked image provides anatomical context. The model learns to synthesize realistic enhancement textures.

**Reward-Guided Training.** We train a scar segmentation model on all real data (including positive and negative LGE images) and incorporate it as a fixed reward model $\mathcal{S}$ (similar to [17], frozen unlike [22]) to provide additional supervision. For each generated image $I_{\text{gen}}$, a predicted mask $M_{\text{gen}} = \mathcal{S}(I_{\text{gen}})$ is compared to the ground truth (GT) $M_{\text{true}}$ using cross-entropy loss. The total training loss combines the diffusion loss and reward loss, weighted by $\lambda_{\text{reward}}$. To reduce computational cost, reward supervision is applied only at early diffusion steps ($t \leq t_{\text{thresh}}$), where the generated image retains meaningful structure. Thus,

$$\mathcal{L}_{\text{total}} = \begin{cases} \mathcal{L}_{\text{diffusion}} + \lambda_{reward} * \mathcal{L}_{\text{reward}} & t <= t_{thresh} \\ \mathcal{L}_{\text{diffusion}} & t > t_{thresh} \end{cases}$$

This one-step reward optimization strategy allows us to inject task-specific supervision into the generative process via the reward model, while avoiding costly multi-step backpropagation. In practice, selecting an effective reward model is challenging, particularly in the domain of LGE segmentation, where large-scale, generalizable models are not readily available. Nonetheless, we demonstrate that even a segmentation model trained on limited real data can provide meaningful supervisory signals that improve generation fidelity and alignment.

**Caption Generation.** For each [LGE image, scar mask] pair, we generate clinically meaningful captions describing scar location and transmurality. Myocardial and RV insertion masks are first obtained using pretrained models, enabling subdivision of the myocardium into 17 AHA segments [1] and three radial layers (endocardial, mid, epicardial). Scar-mask overlap yields anatomical descriptors (e.g., anterior, posteroseptal, endocardial etc.) which are populated into predefined templates to produce captions such as: "Transmural enhancement in the posteroseptal wall". Specific implementation details are given in [9]. This ensures spatially and clinically relevant text conditioning, We compare this against constant-phrase captioning ("LGE image of the heart") in ablation studies.

**Text Encoder.** We replace ControlNet's default encoder (OpenCLIP ViT-H [15]) with BiomedBERT [18], a biomedical language model pretrained on medical texts, to better interpret anatomical descriptors in captions. Its impact is evaluated via ablations. To encourage spatial reliance, 50% of captions are randomly dropped during training [28].

**Synthetic Data Generation.** We train on positive LGE images and synthesize enhancement on negative LGE images using systemically generated conditioning inputs: a scar mask (or edge map) and related caption. As described earlier, for each image, we define the AHA 17-segment and radial layer map, select a random region (e.g., anterolateral, endocardial), and place an ellipsoid, simulating a scar mask. A matching caption is generated as in the previous section. More details are given in [9]. This process enables controlled synthesis of diverse scar patterns, including rare cases.We note that while we only use negative LGE images for inference in this study for simplicity, theoretically the method is not limited to the same, and can be applied to any LGE image. However in such cases, existing LGE patterns need to be taken into account during text and mask generation.

## 4 Experiments

### 4.1 Dataset

The dataset included 212 patients, split into 159 (1516 images) for training (20% validation split) and 53 (497 images) for testing. Scans were acquired on 1.5T scanners (MAGNETOM Aera, Siemens Healthineers) using a T1-weighted inversion-recovery gradient-echo sequence. Expert readers annotated each case using the AHA 17-segment model, assigning binary labels to each segment (1 = scar present, 0 = no scar). Pixel-wise GT was then derived semi-automatically from these labels. Initial masks were generated using the n-standard deviation method [3] with $n = 1.5$, favoring sensitivity. These were manually refined by removing clusters in clinically normal segments, reducing false positives (FP) while preserving true scar regions.

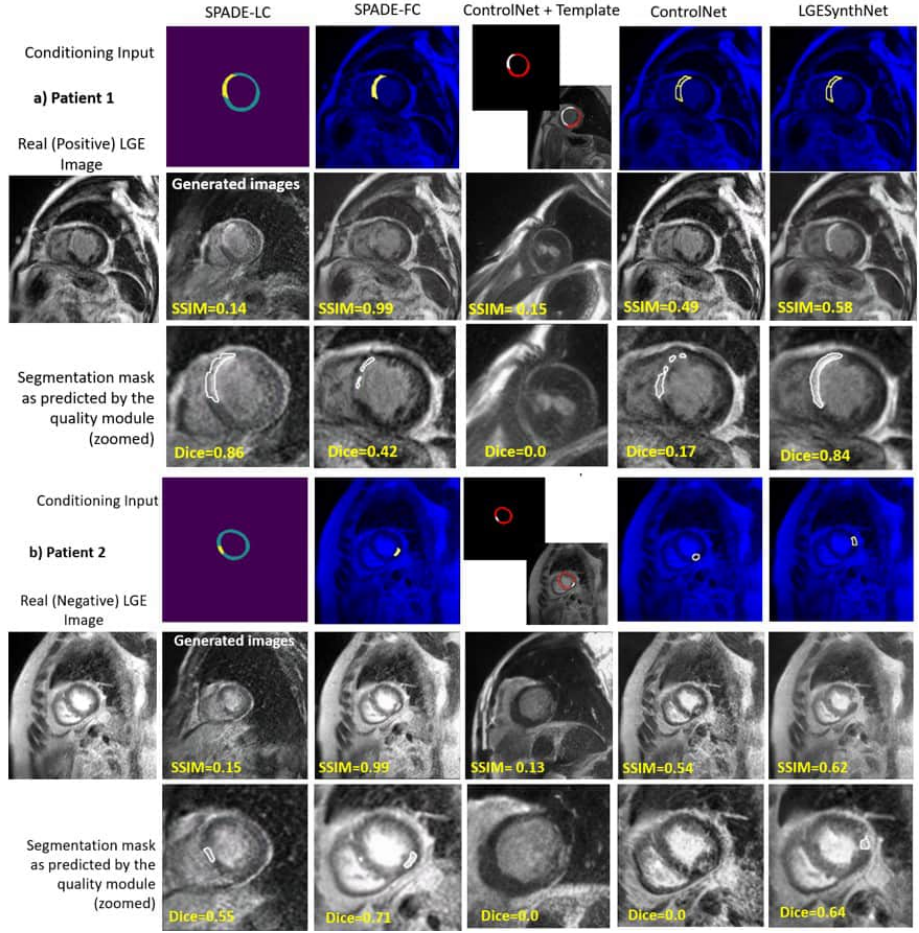

**Fig. 2.** Examples of generated images from each method for two patients. Patient (a) has LGE, and the task is to reconstruct the LGE. Patient (b) does not have LGE, and the task is to construct synthetic LGE. The columns show the results of each method. For each patient, we show the conditioning inputs (row 1), generated images (row 2) and segmentation mask as predicted by the quality module (row 3). The GAN-based SPADE methods and diffusion-based ControlNet methods are explained in Section 4.2 . SSIM (indicating generated synthetic image quality) and Dice coefficient (indicating condition adherence) are shown inlaid on the images in rows 2 and 3.

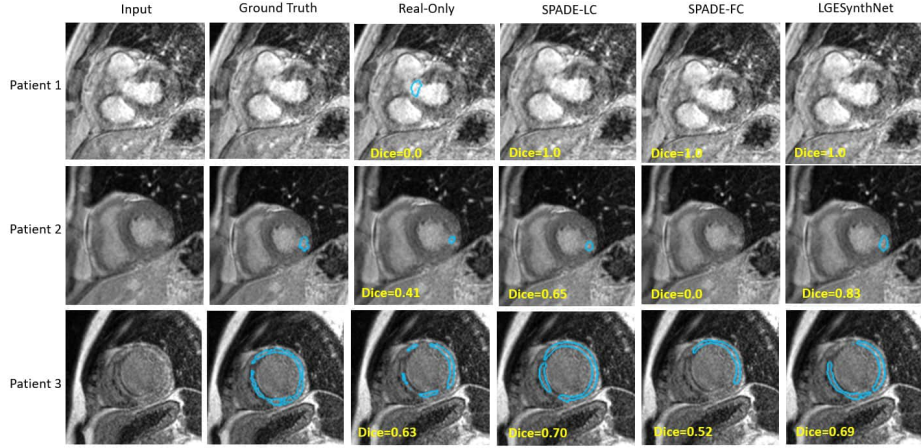

**Fig. 3.** Examples from downstream task (zoomed in, predicted scar mask in in blue) for 3 patients. Model training is augmented with 300 synthetic images generated from 3 methods - SPACE-LC, SPADE-FC and LGESynthNet.

### 4.2   Compared Methods

1. ControlNet: We use the official ControlNet implementation [28], conditioned on scar edge maps and masked LGE images.
2. SPADE [21]: A GAN-based model with strong medical image generation performance [25, 21, 13]. We evaluate two variants: (a) SPADE-LC (Limited Context), conditioned on myocardial and scar masks only (similar to its original design); and (b) SPADE-FC (Full Context), conditioned on scar masks and masked LGE images, paralleling DM inputs.
3. Template Guidance: Adapted from Template Guided Conditional Diffusion Model (TCDM) [25], this strategy appends conditioning inputs with a reference image-mask pair to preserve fine anatomical details and improve structural consistency. We implement this using myocardial and scar masks (query mask) plus a reference consisting of a randomly chosen [LGE image, myocardial mask, scar mask].

**Implementation Details.** All DMs include the caption generation and biomedical text encoder modules. They were trained for 200 epochs, using the final epoch weights for inference with 10 timesteps and classifier-free guidance scale set to 9. Only the ControlNet module was trained; other components were frozen. The reward-guided model used $t_{\text{thresh}} = 200$ and $\lambda_{\text{reward}} = 1$. Edge vs. mask conditioning is analyzed in Table 3b. SPADE models were trained with default parameters. Training used only positive LGE images (patients): 429 (79) for training, 59 (17) for validation, and 149 (29) for testing. Inputs are resampled to 1mm isotropic, cropped to $256 \times 256$, clipped at the 98th percentile, and normalized to $[0, 1]$. Models were trained on 4 A100 40GB GPUs (BS 64).

**Table 1.** Results on test set (n = 79 patients, 149 images, positive LGE only) both image quality and conditioning alignment. All diffusion models (DM) use Caption Generation module and BiomedBERT text encoder.

| Experiment | Conditioning Input Format | Metrics - Image quality | | | | Metrics - Conditioning | |
|---|---|---|---|---|---|---|---|
| | | Whole Image | | Cropped around LV | | | |
| | | SSIM | RMSE | SSIM | RMSE | Dice-TP only | Pass% |
| Real Images | - | 1.0 | 0.00 | 1.0 | 0.00 | 0.465 ± 0.32 | 35.6% |
| **GAN-Based** | | | | | | | |
| SPADE[21]–LC | Semantic Masks | 0.147 ± 0.02 | 0.088 ± 0.02 | 0.159 ± 0.03 | 0.069 ± 0.03 | **0.362 ± 0.30** | **26.2%** |
| SPADE – FC | Semantic Masks + Image Context | **0.974 ± 0.04** | **0.002 ± 0.01** | **0.986 ± 0.01** | **0.001 ± 0.00** | 0.272 ± 0.30 | 18.8% |
| **DM-Based** | | | | | | | |
| ControlNet[28] | Semantic edges + Image Context | 0.571 ± 0.05 | **0.008 ± 0.003** | 0.570 ± 0.06 | 0.009 ± 0.00 | 0.230 ± 0.28 | 12.1% |
| ControlNet + Template[25] | Semantic edges + Reference pair | 0.130 ± 0.03 | 0.102 ± 0.02 | 0.100 ± 0.03 | 0.093 ± 0.03 | 0.002 ± 0.01 | 0.7 % |
| LGESynthNet | Semantic edges + Image Context | **0.587 ± 0.05** | 0.009 ± 0.00 | **0.582 ± 0.06** | 0.009 ± 0.00 | **0.273 ± 0.30** | **18.1%** |

### 4.3 Downstream Task

We evaluate the benefit of synthetic data for scar segmentation. Synthetic images are generated from negative LGE images conditioned on parametrically defined scar masks. To ensure data quality, the reward model is re-used as a quality module to predict scar masks on generated images; only samples with Dice overlap > 0.6 against the conditioning mask are retained. This filter assesses scar location and rough shape but not texture quality. For these selected samples, the predicted mask—not the original ellipsoidal mask—is used as GT for downstream training to better capture realistic scar boundaries.

**Implementation**: We use a DenseNet121 [7] backbone (5 down-sampling levels) to jointly predict scar and myocardium. The model is trained for 200 epochs with Jaccard loss, geometric and elastic augmentations on 4 A100 40 GB GPUs (BS 128), and cosine annealing(initial LR 0.001). Best model is selected via validation Dice, sensitivity, and specificity. For hybrid training, $n = 300$ synthetic samples per generative model are added. Ablations vary N.

**Table 2.** Results on test set: n = 53 patients, 497 real images + 300 synthetic images.

| Experiment | Dice | Dice,TP-only | Accuracy | Bal. Accuracy | Confusion Mat. |
|---|---|---|---|---|---|
| Real Images only | 0.72 | 0.32 | 0.77 | 0.73 | [11, 11, 1, 30] |
| **Trained with real and synthetic images:** | | | | | |
| SPADE – LC | **0.77** | 0.31 | 0.85 | 0.86 | [20, 2, 6, 25] |
| SPADE – FC | 0.75 | 0.19 | 0.68 | 0.72 | [21, 1, 16, 15] |
| LGESynthNet | **0.77** | **0.35** | **0.89** | **0.90** | **[21, 1, 5, 26]** |

### 4.4 Evaluation Metrics

Image quality is assessed using Structural Similarity Index (SSIM) and Root Mean Square Error (RMSE), computed over the entire image and a cropped region of interest around the myocardium. Conditioning consistency is evaluated using the quality control module by reporting the average Dice score and the percentage of samples with Dice $> 0.6$. For the downstream task, we report segmentation Dice score, Dice on True Positive (TP) images-only, and patient-level LGE detection metrics: accuracy, balanced accuracy, and confusion matrix.

## 5   Results

Table 1 shows the results for image quality and conditioning adherence. SPADE-FC yields the best image quality but poor conditioning consistency, often reproducing input images without realistic scar synthesis (also demonstrated in Figure 2). This could be because the method learns pixel-wise normalization factors, which allows it to reproduce the input condition faithfully. SPADE-LC creates more plausible scars but with incoherent backgrounds. The template-guided DM struggles to leverage reference pair for anatomical context, while the reward-supervised LGESynthNet achieves a good balance of fidelity and conditioning, and is chosen for downstream tasks. There is some loss in image quality, as the image information is compressed through the latent space, resulting in lower SSIM and RMSE relative to SPADE-FC. Both SPADE variants are also evaluated downstream for their strengths in conditioning and realism. All methods show low pass rates, indicating challenges in condition adherence.

Table 2 reports downstream segmentation results with synthetic augmentation. SPADE-FC reduces performance despite high image quality, possibly due to unrealistic scar representations. SPADE-LC improves detection but not Dice. LGESynthNet achieves the highest gains, improving Dice by 5 points and patient accuracy by 12–17 points. Figure 3 shows some representative examples. Performance improves with more synthetic samples (Table 3a), stabilizing at N=1000. Ablations (Table 3b) show edge inputs outperform masks, the usage of biomedical encoder and descriptive captions improves performance.

Table 3. Further Analyses on LGESynthNet: a) Effect of number of samples in downstream training. Confusion matrix is given as [TN, FP, FN, TP] b) Ablation studies altering the input condition, text encoder and text generation.

| a) Effect of increasing synthetic samples in downstream training | | | | | |
|---|---|---|---|---|---|
| N samples | Dice | Dice, TP-only | Accuracy | Bal. Accuracy | Confusion Mat. |
| 500 | **0.78** | 0.36 | 0.91 | 0.91 | [21, 1, 4, 27] |
| 1000 | 0.77 | 0.36 | 0.92 | 0.93 | [21, 1, 3, 28] |
| 1500 | 0.77 | **0.37** | **0.91** | **0.91** | **[20, 2, 3, 28]** |

| b) Ablation Analyses | | | | |
|---|---|---|---|---|
| Changes | Whole Image | | Cropped Image | |
| | SSIM | RMSE | SSIM | RMSE |
| [x] Input condition: semantic masks, initialized from SD1.5+semantic masks | 0.497 ± 0.05 | 0.013 ± 0.005 | 0.490 ± 0.07 | 0.013 ± 0.005 |
| [x] Text encoder: OpenCLIP CLIP-ViT-H [15] | 0.554 ± 0.05 | 0.011 ± 0.007 | 0.550 ± 0.06 | 0.011 ± 0.006 |
| [x] Constant text caption | 0.580 ± 0.05 | 0.008 ± 0.003 | 0.572 ±0.07 | 0.009 ± 0.004 |

## 6  Discussion

We present LGESynthNet, a latent DM-based framework for generating controllable, high-quality synthetic LGE images using limited annotated data. Framing the task as inpainting reduces generation complexity, while reward modeling and domain-specific design choices improve conditioning fidelity. A quality control module enables selective use of reliable synthetic samples, addressing variability in conditioning adherence. Our results highlight the need to assess generation across three dimensions: a) image realism, b) conditioning consistency, and c) downstream utility. Importantly, high image quality does not imply alignment to conditioning, and alignment is difficult to quantify (location, texture etc.), making downstream performance a critical benchmark. Limitations include the use of single-center data and semi-automated GT masks based on segment-level labels. Future work includes evaluation of generalization of the model and expanding the data to multi-center, multi-vendor cohorts. Additionally, in this work, we have prompted the DM with simple, elliptical scar shapes. Future work involves exploring more realistic clinical patterns of LGE occurrence.

**Disclosure of Interests.** The authors AJJ and PS are employees of Siemens Healthineers.
**Disclaimer.** The concepts and information presented in this paper are based on research results that are not commercially available. Future commercial availability cannot be guaranteed.